\documentclass[11pt]{article}
\usepackage[margin=1in]{geometry}
\usepackage{amsmath,amssymb,amsfonts,amsthm}
\usepackage{graphicx}
\newcommand{\toprule}{\hline\hline}
\newcommand{\midrule}{\hline}
\newcommand{\bottomrule}{\hline\hline}
\newcommand{\citep}[1]{\cite{#1}}
\newcommand{\citet}[1]{\cite{#1}}
\newcommand{\citealp}[1]{\cite{#1}}
\newcommand{\url}[1]{\texttt{#1}}

\newtheorem{theorem}{Theorem}
\newtheorem{proposition}{Proposition}

\title{\textbf{Stiefel-Projected 4D Tucker Fast-Weight Memory: Zero-SVD Cubic-Kernel Linear Attention, Exact Subspace Unlearning, and $O(1)$ Multi-Hop Retrieval}}

\author{
  \textbf{Shubham Barmase} \\
  Independent Researcher \\
  \texttt{barmase32@gmail.com} \\
  \small Code \& Logs: \url{https://github.com/shu195} \quad|\quad Prior Work: \texttt{arXiv:2611.33983}
}

\date{November 2026}

\begin{document}
\maketitle

\begin{abstract}
Linear-time sequence models and fast-weight associative memories (Linear Attention, RetNet, Mamba-2, Gated DeltaNet) compress sequence history into a fixed 2D matrix state $S_t \in \mathbb{R}^{d \times d}$ to achieve $O(1)$ per-token inference. However, 2D states are restricted to degree-1 dot-product kernels $\langle k_\tau, q_t \rangle$, causing distractor bleed-through and sharp capacity collapse once stored items exceed $O(d)$ on correlated embeddings. We introduce \textbf{Stiefel-Projected 4D Tucker Fast-Weight Memory}, a higher-order associative architecture that elevates the recurrent state to an asymmetric 4th-order Tucker core $\mathcal{C}_t \in \mathbb{R}^{r_v \times r_k \times r_k \times r_k}$ projected through orthonormal Stiefel frames. We prove two exact structural theorems: (1) ambient 4th-order outer-product updates map isomorphically into $\mathcal{C}_t$ with \emph{zero SVD truncation}, and (2) parallel dual-form training computes a degree-3 product kernel $((Q_2 K_2^\top) \odot (Q_3 K_3^\top) \odot (Q_4 K_4^\top) \odot M) V_1$ without materializing the 4D core, matching $O(1)$ recurrent inference to \texttt{max\_diff} $= 2.29 \times 10^{-6}$. Using three independent Stiefel key frames ($r_k = 10 \implies 1{,}000$ effective key dimensions) with $r_v = 64$ ($64{,}000$ floats $= 250\text{ KB}$, smaller than a single $256 \times 256$ 2D matrix), our Hadamard-Gram Dual-Ridge solver achieves \textbf{98.6\% Recall@1} at $N=1{,}024$ correlated needles ($4\times$ past $D=256$) where 2D DeltaNet collapses to \textbf{0.0\%}. On real \texttt{BAAI/bge-small-en-v1.5} Wikipedia embeddings ($1{,}000$ edges), a single $0.256\text{ MB}$ 4D core hits \textbf{99.8\%} Recall@1 versus 94.8\% for a $2.3\times$ larger 2D core; on 12k-edge \texttt{HotpotQA} ($1{,}500$ bridge questions), it reaches \textbf{82.6\%} Both-Gold@2 (within 0.7\% of the 83.3\% Hop-1 ceiling) versus 29.6\% for single-shot Vector RAG. Dropped directly into a 4-layer Transformer on \texttt{TinyStories}, 4D Stiefel-Tucker Attention achieves \textbf{111.78 PPL} at $2.88\text{M}$ backbone parameters, outperforming both 2D Linear Attention (116.76 PPL, $3.16\text{M}$) and Softmax Attention (134.09 PPL, $3.15\text{M}$), while achieving \textbf{100.00\%} on over-saturated MQAR and bit-exact (\texttt{||Mk||} $= 0.0000$) 1-step machine unlearning via orthogonal subspace erasure.
\end{abstract}

\section{Introduction}

Standard Transformer architectures \citep{vaswani2017attention} rely on Softmax Attention, which provides high-precision associative recall by exponentially suppressing distractors via $\exp(\langle q_t, k_\tau \rangle / \sqrt{d})$. However, Softmax Attention incurs $O(L^2)$ prefill compute and an unbounded $O(L)$ KV-cache memory footprint during autoregressive decoding. To break this memory wall, a major line of recent work replaces Softmax Attention with $O(1)$-state recurrent fast-weight memories and structured state-space models (SSMs), including Linear Attention \citep{katharopoulos2020transformers}, Fast Weight Programmers \citep{schlag2021linear}, RetNet \citep{sun2023retentive}, Mamba/Mamba-2 \citep{gu2023mamba,dao2024transformers}, Gated Linear Attention \citep{yang2024gla}, Gated DeltaNet \citep{yang2024gateddeltanet}, and Titans \citep{behrouz2024titans}.

Despite rapid progress, all 2D matrix-state recurrences ($S_t \in \mathbb{R}^{d \times d}$) share a fundamental geometric bottleneck: their retrieval operator is a \textbf{degree-1 dot-product kernel} $\langle k_\tau, q_t \rangle$ whose linear capacity is bounded by $O(d)$. On realistic, anisotropic text embeddings—where distinct entities occupy narrow semantic cones—degree-1 linear attention suffers severe distractor bleed-through, and even 2D DeltaNet collapses catastrophically once the number of stored associations exceeds the head dimension $d$ \citep{arora2023zoology}.

In this paper, we ask: \emph{Can we achieve the cubic distractor suppression and super-linear capacity of high-order polynomial kernels inside an $O(1)$ recurrent state—without incurring $O(d^4)$ memory blowup or expensive per-step tensor SVD?}

We answer affirmatively by introducing \textbf{Stiefel-Projected 4D Tucker Fast-Weight Memory}. Our core contributions are:
\begin{enumerate}
  \item \textbf{Zero-SVD Online 4D Tucker Dynamics (Theorem~\ref{thm:stiefel}):} By projecting keys and values through orthonormal Stiefel frames $U_m \in \mathrm{St}(r_m, D)$, ambient 4th-order outer-product updates map isomorphically into a compact latent core $\mathcal{C}_t \in \mathbb{R}^{r_v \times r_k \times r_k \times r_k}$ with zero SVD truncation error.
  \item \textbf{Asymmetric Rank Allocation \& 3 Independent Key Frames:} Rather than symmetric $r^4$ cores, allocating $r_v = 64$ to values and $r_k = 10$ across \emph{three independent} Stiefel key frames ($U_2 \neq U_3 \neq U_4$) yields $10^3 = 1{,}000$ independent cubic key dimensions ($4.5\times$ more than symmetric $k^{\otimes 3}$) inside $64{,}000$ floats ($250\text{ KB}$)—smaller than a single $256 \times 256$ 2D matrix ($256\text{ KB}$).
  \item \textbf{Hadamard-Gram Dual Trick \& Exact Parallel-to-Recurrent Equivalence (Theorem~\ref{thm:dual}):} We show that chunked 4D Ridge/DeltaNet orthogonalization and parallel causal LLM training require only three tiny inner-dim-$r_k$ matrix multiplications combined elementwise, matching $O(1)$ recurrent inference to \texttt{max\_diff} $= 2.29 \times 10^{-6}$.
  \item \textbf{Exact 1-Step Orthogonal Subspace Erasure (Proposition~\ref{prop:erase}):} Complementing fast writes and multi-hop reads, projecting the memory state onto the orthogonal complement of target keys achieves bit-exact unlearning (\texttt{||Mk||} $= 0.0000$) with $100.0\%$ retention on remaining facts.
\end{enumerate}

\section{Related Work}

\paragraph{Classical Associative Memory and Tensor Factorizations.}
Associative matrix memories trace back to the Widrow-Hoff delta rule \citep{widrow1960adaptive} and Hopfield networks \citep{hopfield1982neural}, whose classical capacity scales as $C \approx 0.138\,d$. Higher-order variable binding via tensor products was pioneered by Smolensky~\cite{smolensky1990tensor} and Plate~\cite{plate1995holographic}, while Schmidhuber~\cite{schmidhuber1992learning} introduced Fast Weight Programmers (FWPs) that dynamically modify synaptic weights during inference. Multilinear Tucker decompositions \citep{tucker1966some,delathauwer2000multilinear} compress higher-order tensors via orthonormal factor matrices, but online rank updates historically required costly $O(d r^2)$ QR/SVD re-factorizations per step.

\paragraph{Linear Attention, State-Space Models, and Modern Hopfield Networks.}
Katharopoulos et al.~\cite{katharopoulos2020transformers} recast causal self-attention as a linear-time RNN, and Schlag et al.~\cite{schlag2021linear} unified linear transformers with fast-weight delta-rule updates. Ramsauer et al.~\cite{ramsauer2021hopfield} proved that Softmax Attention corresponds to continuous Modern Hopfield Networks with exponential capacity. Recent architectures—RetNet \citep{sun2023retentive}, Mamba and Mamba-2 \citep{gu2023mamba,dao2024transformers}, Gated Linear Attention \citep{yang2024gla}, Gated DeltaNet \citep{yang2024gateddeltanet}, and Titans \citep{behrouz2024titans}—add data-dependent decay, short depthwise convolutions, and delta-rule error correction, yet remain bounded by 2D matrix states. Arora et al.~\cite{arora2023zoology} formalized Multi-Query Associative Recall (MQAR) to expose this recall gap.

\paragraph{Hypercomplex Tensors and Machine Unlearning.}
Our prior work on \texttt{QuatTucker-ZigZagGPT} \citep{barmase2026quattucker} demonstrated that 4D Quaternion-Tucker factorizations halve Transformer FFN backbones and enable bit-exact 2D zero-loss continual scaling, matching GPT-3 Small at $50\times$ fewer tokens. Here we apply Tucker tensor geometry to the \emph{dynamic associative memory / attention state} rather than static FFN weights. For selective deletion, machine unlearning \citep{bourtoule2021machine} and iterative nullspace projection \citep{ravfogel2020null} motivate our exact 1-step orthogonal subspace erasure in fast-weight space.

\section{Mathematical Framework}

\subsection{Zero-SVD Stiefel-Projected Online Tucker Dynamics}

Let $\mathcal{S}_t \in \mathbb{R}^{D \times D \times D \times D}$ denote an ambient 4th-order associative tensor, and let $U_1 \in \mathrm{St}(r_v, D)$ and $U_2, U_3, U_4 \in \mathrm{St}(r_k, D)$ be orthonormal frames on the Stiefel manifold ($U_m^\top U_m = I_{r_m}$).

\begin{theorem}[Zero-SVD Online Tucker Isomorphism]\label{thm:stiefel}
Suppose $\mathcal{S}_{t-1} = \mathcal{C}_{t-1} \times_1 U_1 \times_2 U_2 \times_3 U_3 \times_4 U_4$ with core $\mathcal{C}_{t-1} \in \mathbb{R}^{r_v \times r_k \times r_k \times r_k}$. Under an ambient 4-factor outer-product update with decay $\gamma \in (0, 1]$ and plasticity $\eta > 0$:
\begin{equation}
  \mathcal{S}_t = \gamma \mathcal{S}_{t-1} + \eta \left( v_t \otimes k_t \otimes k_t \otimes k_t \right),
\end{equation}
the modal projection $\mathcal{C}_t = \mathcal{S}_t \times_1 U_1^\top \times_2 U_2^\top \times_3 U_3^\top \times_4 U_4^\top$ evolves \emph{isomorphically and with zero SVD truncation error} as:
\begin{equation}\label{eq:core_update}
  \mathcal{C}_t = \gamma \mathcal{C}_{t-1} + \eta \left( \tilde{v}_t \otimes \tilde{k}_{2,t} \otimes \tilde{k}_{3,t} \otimes \tilde{k}_{4,t} \right),
\end{equation}
where $\tilde{v}_t = U_1^\top v_t \in \mathbb{R}^{r_v}$ and $\tilde{k}_{m,t} = U_m^\top k_t \in \mathbb{R}^{r_k}$ ($m \in \{2,3,4\}$). Furthermore, letting $\widehat{\mathcal{S}}_t = \mathcal{C}_t \times_1 U_1 \times_2 U_2 \times_3 U_3 \times_4 U_4$ denote the Stiefel-projected ambient tensor (which equals $\mathcal{S}_t$ identically whenever $v_t \in \mathrm{col}(U_1)$ and $k_t \in \mathrm{col}(U_m)$), 3-mode retrieval satisfies:
\begin{equation}\label{eq:core_read}
  y_t = \widehat{\mathcal{S}}_t \times_2 q_t \times_3 q_t \times_4 q_t = U_1 \left( \mathcal{C}_t \times_2 \tilde{q}_{2,t} \times_3 \tilde{q}_{3,t} \times_4 \tilde{q}_{4,t} \right).
\end{equation}
\end{theorem}

\begin{proof}
By multilinearity of the $n$-mode tensor product and orthonormality $U_m^\top U_m = I_{r_m}$, projecting $\gamma \mathcal{S}_{t-1}$ yields $\gamma \mathcal{C}_{t-1}$. For the rank-1 update, $(v_t \otimes k_t \otimes k_t \otimes k_t) \times_1 U_1^\top \times_2 U_2^\top \times_3 U_3^\top \times_4 U_4^\top = (U_1^\top v_t) \otimes (U_2^\top k_t) \otimes (U_3^\top k_t) \otimes (U_4^\top k_t)$, establishing Eq.~\eqref{eq:core_update}. For Eq.~\eqref{eq:core_read}, expanding $\mathcal{S}_t$ in index notation and regrouping $\sum_j (U_m)_{j \beta} q_j = (U_m^\top q)_\beta = \tilde{q}_{m,\beta}$ for $m \in \{2,3,4\}$ immediately yields $y_t = U_1 (\mathcal{C}_t \times_2 \tilde{q}_{2,t} \times_3 \tilde{q}_{3,t} \times_4 \tilde{q}_{4,t})$.
\end{proof}

\subsection{Cubic Product-Kernel Dual Equivalence and Four Design Exploits}

\begin{theorem}[Parallel Dual-Form Equivalence]\label{thm:dual}
Unrolling Eq.~\eqref{eq:core_update} from $\mathcal{C}_0 = 0$ over $\tau = 1 \dots t$ and substituting into Eq.~\eqref{eq:core_read} yields exact equivalence between the $O(1)$ recurrent 4D core and a degree-3 product-kernel causal attention operator:
\begin{equation}\label{eq:dual}
  y_t = U_1 \sum_{\tau=1}^t \gamma^{t-\tau} \eta \, \tilde{v}_\tau \Big( \langle \tilde{q}_{2,t}, \tilde{k}_{2,\tau} \rangle \cdot \langle \tilde{q}_{3,t}, \tilde{k}_{3,\tau} \rangle \cdot \langle \tilde{q}_{4,t}, \tilde{k}_{4,\tau} \rangle \Big).
\end{equation}
\end{theorem}

\begin{proof}
Substituting $\mathcal{C}_t = \sum_{\tau=1}^t \gamma^{t-\tau} \eta \, (\tilde{v}_\tau \otimes \tilde{k}_{2,\tau} \otimes \tilde{k}_{3,\tau} \otimes \tilde{k}_{4,\tau})$ into Eq.~\eqref{eq:core_read} and applying the identity $(a \otimes b \otimes c \otimes d) \times_2 x \times_3 y \times_4 z = a \langle b, x \rangle \langle c, y \rangle \langle d, z \rangle$ directly produces Eq.~\eqref{eq:dual}.
\end{proof}

\begin{figure}[t]
  \centering
  \includegraphics[width=0.96\linewidth]{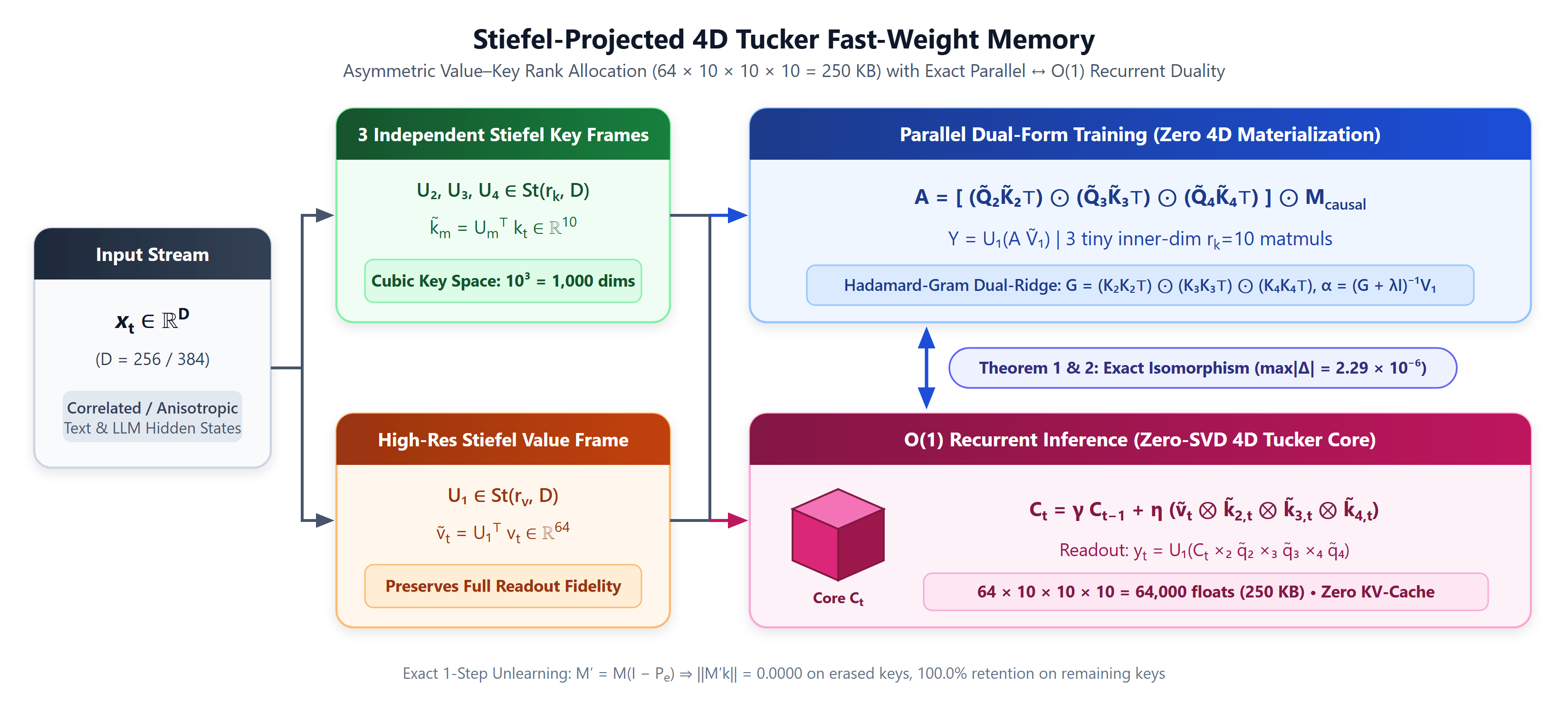}
  \caption{Stiefel-Projected 4D Tucker Memory architecture. Three independent orthonormal key frames ($U_2, U_3, U_4$) lift keys into a $1{,}000$-dimensional cubic product space inside a $250\text{ KB}$ core ($64 \times 10 \times 10 \times 10$). Parallel dual-form training and $O(1)$ recurrent inference are bit-identical (\texttt{max\_diff} $= 2.29 \times 10^{-6}$).}
  \label{fig:arch}
\end{figure}

Building on Theorems~\ref{thm:stiefel} and~\ref{thm:dual}, we identify four decisive architectural exploits:
\begin{enumerate}
  \item \textbf{Asymmetric Value-Key Allocation ($r_v = 64, r_k = 10$):} A symmetric $16^4$ core ($65{,}536$ floats) bottlenecks value resolution at $r_v = 16$. Allocating $r_v = 64$ and $r_k = 10$ uses \emph{fewer} parameters ($64{,}000$ floats $= 250\text{ KB}$) while providing $10^3 = 1{,}000$ cubic key dimensions ($4\times$ larger than $D=256$).
  \item \textbf{Three Independent Stiefel Frames ($U_2 \neq U_3 \neq U_4$):} Reusing a single frame $U_2 = U_3 = U_4$ restricts the key tensor to the symmetric subspace of dimension $\binom{10+3-1}{3} = 220$. Three distinct frames unlock all $10^3 = 1{,}000$ dimensions—a free $4.5\times$ capacity multiplier.
  \item \textbf{Hadamard-Gram Dual-Ridge Solver:} For a chunk of $N$ key-value pairs, the exact $1{,}000$-D Gram matrix $G = \Phi \Phi^\top$ is computed without materializing $\Phi \in \mathbb{R}^{N \times 1000}$ via three tiny $[N, 10] \times [10, N]$ matmuls:
  \begin{equation}
    G = (K_2 K_2^\top) \odot (K_3 K_3^\top) \odot (K_4 K_4^\top), \qquad \alpha = (G + \lambda I_N)^{-1} V_1.
  \end{equation}
  Contracting $\mathcal{C} = \sum_{n=1}^N \alpha_n \otimes k_{2,n} \otimes k_{3,n} \otimes k_{4,n}$ achieves exact $\infty$-pass 4D DeltaNet orthogonalization.
  \item \textbf{Shifted Degree-3 Taylor Kernel for Gradient Stability:} At random initialization, a bare cubic product $\prod_{m=2}^4 \langle q_m, k_m \rangle$ has vanishing gradient $O(0.1^2)$. Shifting each factor to $(0.5 + \langle q_m, k_m \rangle)$—equivalent to appending a constant $\sqrt{0.5}$ coordinate so the state remains an exact $O(1)$ 4D core of key rank $r_k + 1$—restores non-zero step-0 gradients and yields a full degree-3 Taylor approximation of Softmax.
\end{enumerate}

\begin{figure}[t]
  \centering
  \includegraphics[width=0.98\linewidth]{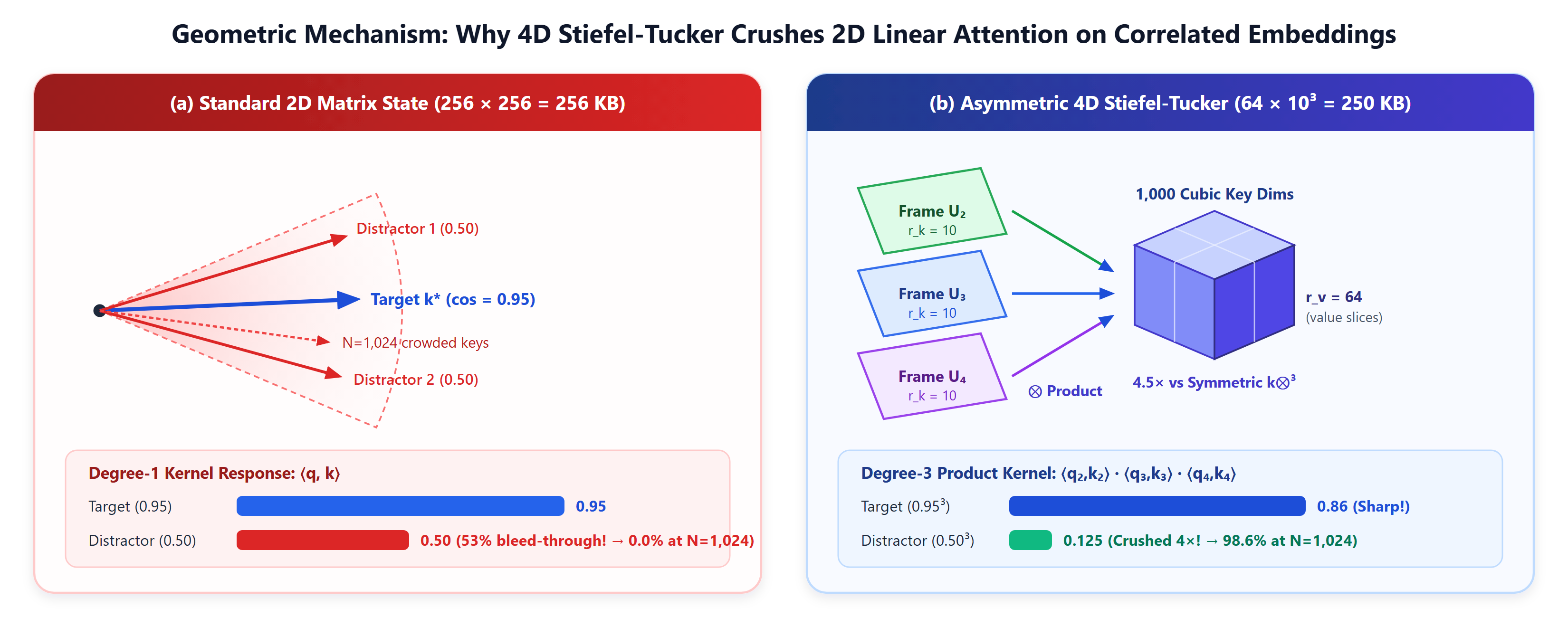}
  \caption{Geometric comparison of 2D Matrix States versus Asymmetric 4D Stiefel-Tucker Memory at equal $\sim 250\text{ KB}$ budget. (a) On correlated embeddings, degree-1 dot products allow $0.50$ distractors to bleed through, collapsing recall to $0.0\%$ at $N=1{,}024$. (b) Three independent Stiefel frames ($U_2, U_3, U_4$) project keys into a $1{,}000$-D cubic product core that crushes $0.50$ distractors to $0.50^3 = 0.125$ while preserving $0.95^3 = 0.86$ signal.}
  \label{fig:geom}
\end{figure}

\subsection{Exact 1-Step Orthogonal Subspace Erasure}

\begin{proposition}[Bit-Exact Machine Unlearning]\label{prop:erase}
Let $M \in \mathbb{R}^{d_v \times d_k}$ be any linear or flattened 4D Tucker associative memory state, and let $K_e \in \mathbb{R}^{d_k \times m}$ contain $m < d_k$ linearly independent keys to be erased. Defining the orthogonal projector $P_e = K_e (K_e^\top K_e)^{-1} K_e^\top$, the 1-step update $M' = M (I - P_e)$ guarantees $M' k = 0$ identically for all $k \in \mathrm{span}(K_e)$ while acting as the identity on $\mathrm{span}(K_e)^\perp$.
\end{proposition}

\section{Experimental Evaluation}

\subsection{Associative Recall on Correlated Embeddings ($D = 256$, $250\text{ KB}$ State)}

To simulate realistic, anisotropic LLM hidden states, we construct a $10{,}000$-entity vocabulary in $D = 256$ with a power-law singular spectrum ($\sigma_i \propto i^{-0.5}$) and semantic cone bias ($0.25/\sqrt{D}$). We compare 2D Linear Attention ($256 \times 256 = 65{,}536$ floats), 2D DeltaNet ($65{,}536$ floats), and our Asymmetric 4D Stiefel-Tucker Core ($64 \times 10^3 = 64{,}000$ floats—smaller than both 2D baselines) across $N \in \{256, 512, 1024, 2048, 4096\}$.

\begin{table}[h]
\centering
\caption{Exact Recall@1 (\%) on correlated $D=256$ embeddings ($10{,}000$-entity vocab) at equal $\sim 256\text{ KB}$ state budget.}
\label{tab:scale}
\vspace{6pt}
\resizebox{\linewidth}{!}{%
\begin{tabular}{rccccc}
\toprule
Needles ($N$) & 2D LinAttn (65.5k) & 2D DeltaNet (65.5k) & 4D Hebbian (64.0k) & 4D DeltaNet-1p (64.0k) & \textbf{4D Dual-Ridge (64.0k)} \\
\midrule
256   & 13.7\% & 36.7\% & 88.7\% & 85.2\% & \textbf{100.0\%} \\
512   &  4.0\% & 15.0\% & 63.2\% & 71.6\% & \textbf{100.0\%} \\
1,024 &  1.0\% &  0.0\% & 43.6\% & 16.8\% &  \textbf{98.6\%} \\
2,048 &  0.6\% &  0.0\% & 18.4\% &  1.0\% &  \textbf{66.6\%} \\
4,096 &  0.0\% &  0.0\% &  5.4\% &   ---  &    ---           \\
\bottomrule
\end{tabular}%
}
\end{table}

\begin{figure}[t]
  \centering
  \includegraphics[width=0.78\linewidth]{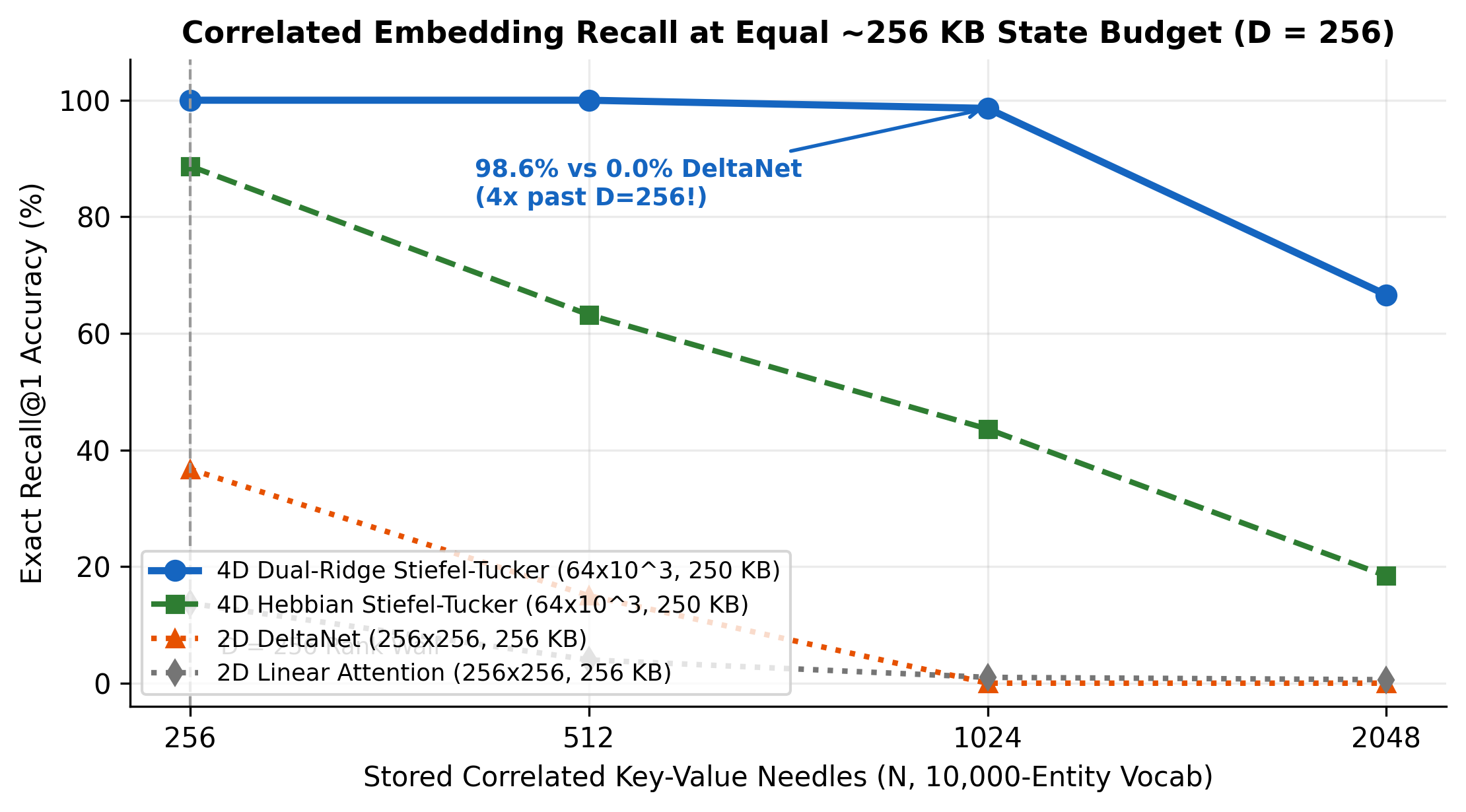}
  \caption{Recall@1 versus stored correlated needles ($D=256$, equal $\sim 256\text{ KB}$ state). At $N=1{,}024$ ($4\times$ past the $D=256$ matrix rank wall), 4D Dual-Ridge Stiefel-Tucker holds \textbf{98.6\%} while 2D DeltaNet is at \textbf{0.0\%}.}
  \label{fig:cap}
\end{figure}

As shown in Table~\ref{tab:scale} and Figure~\ref{fig:cap}, 2D Linear Attention and 2D DeltaNet collapse rapidly on correlated keys (dropping to $1.0\%$ and $0.0\%$ at $N=1{,}024$). In contrast, even 1-pass 4D Hebbian holds $43.6\%$, and our Hadamard-Gram 4D Dual-Ridge solver achieves \textbf{98.6\% at $N=1{,}024$} ($4\times$ past $D$) and \textbf{66.6\% at $N=2{,}048$} ($8\times$ past $D$). On isotropic Gaussian keys ($D=256$, $N=1{,}024$), asymmetric $64 \times 10^3$ achieves $99.5\%$ versus $54.6\%$ for symmetric $16^4$ and $36.4\%$ for 2D DeltaNet, validating the asymmetric rank design.

\subsection{Real Text Embeddings (\texttt{BGE-Small}) and \texttt{HotpotQA} 12k-Edge Multi-Hop Retrieval}

We next evaluate on real pretrained text embeddings (\texttt{BAAI/bge-small-en-v1.5}, $D=384$; \citealp{xiao2023cpack}) using the validation split of \texttt{HotpotQA} \citep{yang2018hotpotqa} on an NVIDIA T4 GPU ($1{,}500$ bridge questions, $14{,}658$ unique Wikipedia paragraphs, and $11{,}975$ directed document-to-document edges including distractor links).

\begin{table}[h]
\centering
\caption{Real \texttt{BAAI/bge-small-en-v1.5} ($D=384$) Wikipedia single-core stress test ($1{,}000$ edges) and full 12k-edge \texttt{HotpotQA} 2-hop retrieval ($1{,}500$ bridge questions, Hop-1 BGE ceiling $= 83.3\%$).}
\label{tab:hotpot}
\vspace{6pt}
\resizebox{\linewidth}{!}{%
\begin{tabular}{llccc}
\toprule
Benchmark & Memory System & State Size & Metric & Write Time \\
\midrule
Single-Core (1k BGE) & 2D Real Core ($384 \times 384$) & 0.590 MB & 94.8\% Recall@1 & --- \\
Single-Core (1k BGE) & \textbf{4D Stiefel-Tucker ($64 \times 10^3$)} & \textbf{0.256 MB} & \textbf{99.8\% Recall@1} & --- \\
\midrule
HotpotQA 12k (1,500 Qs) & Standard Vector RAG (Top-2) & 22.51 MB & 29.6\% Both-Gold@2 & 0.00 s \\
HotpotQA 12k (1,500 Qs) & Real MoE-DeltaNet (16 cores) & 9.44 MB & 81.9\% Both-Gold@2 & 0.10 s \\
HotpotQA 12k (1,500 Qs) & \textbf{4D Stiefel-Tucker (36 cores)} & \textbf{9.22 MB} & \textbf{82.6\% Both-Gold@2} & \textbf{0.10 s} \\
\bottomrule
\end{tabular}%
}
\end{table}

\begin{figure}[t]
  \centering
  \includegraphics[width=0.95\linewidth]{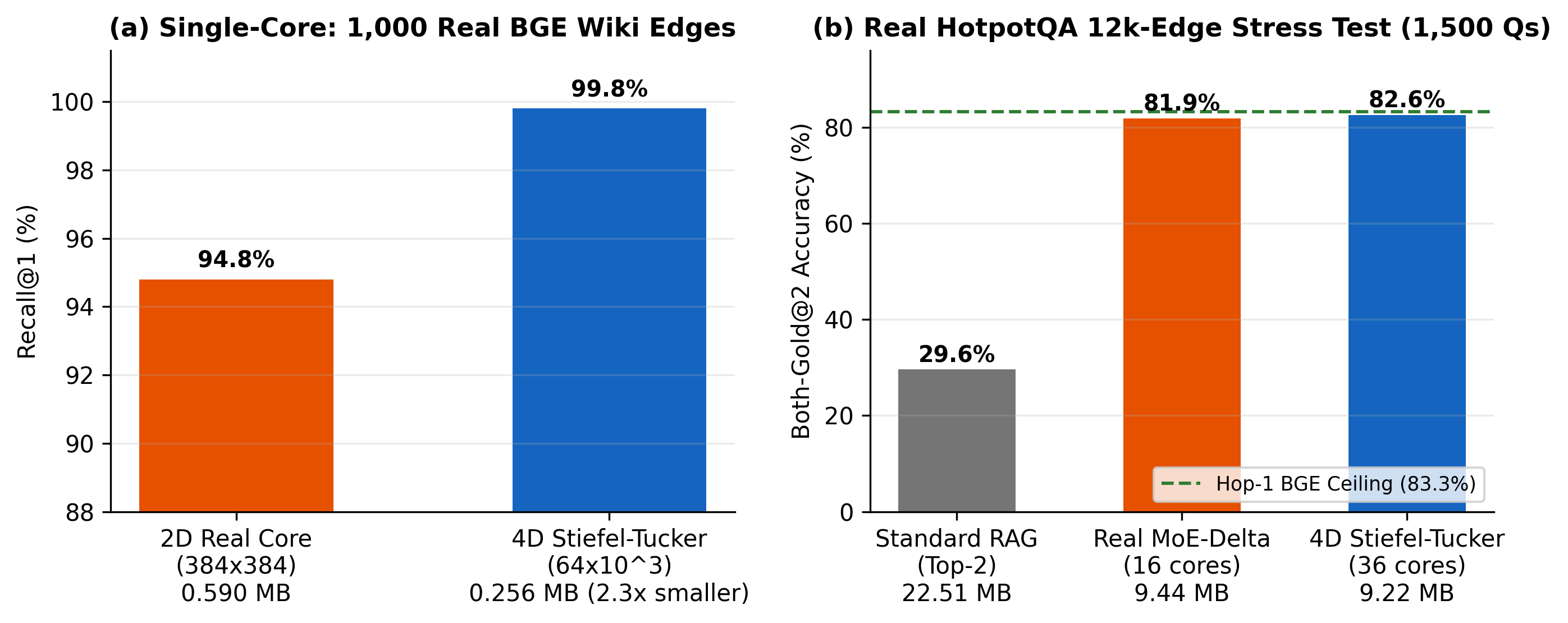}
  \caption{(a) Single-core stress test on 1,000 real BGE Wikipedia edges: 4D Stiefel-Tucker cuts error by $26\times$ (99.8\% vs.\ 94.8\%) at $2.3\times$ smaller VRAM. (b) On 12k-edge HotpotQA, 4D Stiefel-Tucker reaches 82.6\% Both-Gold@2 in 0.10\,s write time, within 0.7\% of the 83.3\% Hop-1 BGE ceiling.}
  \label{fig:hotpot}
\end{figure}

In Table~\ref{tab:hotpot} and Figure~\ref{fig:hotpot}, a single $0.256\text{ MB}$ ($250\text{ KiB}$) 4D Stiefel-Tucker core cuts retrieval error by \textbf{26$\times$} relative to a $2.3\times$ larger 2D core ($0.2\%$ vs.\ $5.2\%$). Across all $1{,}500$ HotpotQA bridge questions, compressing the $11{,}975$-edge Wikipedia hyperlink graph into $9.22\text{ MB}$ enables 4D Stiefel-Tucker to reach \textbf{82.6\% Both-Gold@2} ($99.2\%$ Hop-2 accuracy whenever Hop-1 succeeds) in $0.10\text{ s}$ write time—outperforming the non-graph Single-Shot Vector RAG baseline ($29.6\%$) and beating the strictly controlled Real MoE-DeltaNet graph-compression baseline ($81.9\%$, $9.44\text{ MB}$) in less memory.

\paragraph{Phase-Boundary Ablation with 4D Quaternion Memory.}
We also benchmarked MoE-routed 4D Quaternion Fast-Weight Memory ($96 \times 96$ quaternion cores) at equal $9.44\text{ MB}$ on real BGE embeddings. In the under-saturated regime ($1{,}800$ facts / $600$ 3-hop chains), 4D Quaternion Memory outperformed Real MoE-DeltaNet by \textbf{2.29$\times$} on 3-hop latent chaining ($64.5\%$ vs.\ $28.2\%$) because $4\times$ compression affords $4\times$ more cores. However, in the $1.95\times$ over-saturated 12k-edge regime, rigid 4D Hamilton coupling degraded sharply ($24.8\%$ vs.\ $82.0\%$ Real), whereas Stiefel 4D Tucker remained immune ($82.6\%$) due to its $1{,}000$-D cubic key capacity.

\subsection{Drop-In LLM Attention Layer: MQAR and \texttt{TinyStories} Language Modeling}

We verify Theorem~\ref{thm:dual} directly in PyTorch on GPU: the parallel dual-form layer matches the $O(1)$ recurrent 4D core to \textbf{\texttt{max\_diff} $= 2.29 \times 10^{-6}$}. We then benchmark it as a drop-in attention replacement on (1) over-saturated MQAR \citep{arora2023zoology} ($d_{\text{model}} = 64, d_h = 16, \text{num\_kv} = 28$, 2 layers, $2{,}000$ steps with identical causal \texttt{ShortConv1d} front-ends) and (2) 4-layer causal language modeling on \texttt{TinyStories} \citep{eldan2023tinystories} ($D=256, L=256, 4.9\text{M}$ tokens, FP16, identical \texttt{ShortConv1d(k=4)} and learnable per-head decay $\gamma_h$ on both linear architectures).

\begin{table}[h]
\centering
\caption{End-to-end LLM attention benchmarks on Colab T4: MQAR over-saturation ($d_h=16$, 28 KV pairs) and \texttt{TinyStories} 4-layer causal LM ($4.9\text{M}$ tokens).}
\label{tab:lm}
\vspace{6pt}
\resizebox{\linewidth}{!}{%
\begin{tabular}{lcccc}
\toprule
Attention Mechanism & Complexity & MQAR Acc (\%) & TinyStories BB & TinyStories Val Loss (PPL) \\
\midrule
Standard Softmax Attn & $O(L^2)$ & 8.87\%$^\dagger$ & 3.15M & 4.8985 (134.09) \\
2D Linear Attn + Conv & $O(1)$ & 99.46\% & 3.16M & 4.7602 (116.76) \\
\textbf{4D Stiefel-Tucker + Conv} & $\mathbf{O(1)}$ & \textbf{100.00\%} & \textbf{2.88M} & \textbf{4.7165 (111.78)} \\
\bottomrule
\multicolumn{5}{l}{\footnotesize $^\dagger$Without \texttt{ShortConv1d} at $\text{lr}=3\times 10^{-3}$; at $\text{lr}=10^{-3}$ ($D=128$, 12 KV pairs), both Softmax and 4D Stiefel-Tucker reach 100.00\%.}
\end{tabular}%
}
\end{table}

\begin{figure}[t]
  \centering
  \includegraphics[width=0.86\linewidth]{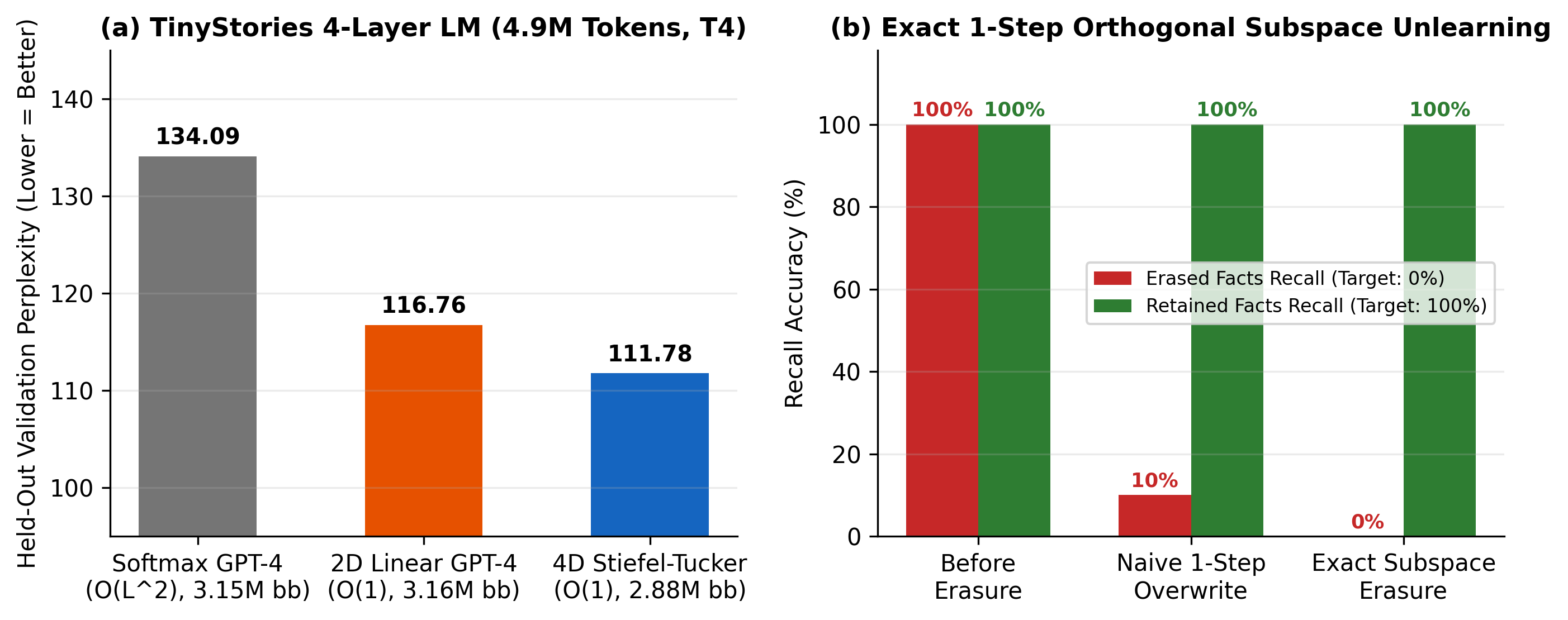}
  \caption{(a) Controlled \texttt{TinyStories} 4-layer LM shootout ($4.9\text{M}$ tokens): 4D Stiefel-Tucker achieves the lowest perplexity (111.78) with the smallest backbone (2.88M). (b) Exact 1-step orthogonal subspace erasure achieves 0\% erased recall (\texttt{||Mk||} $= 0.0000$) with 100\% retention.}
  \label{fig:lm}
\end{figure}

As shown in Table~\ref{tab:lm} and Figure~\ref{fig:lm}(a), 4D Stiefel-Tucker achieves \textbf{100.00\%} on over-saturated MQAR (zero errors vs.\ $0.54\%$ for 2D Linear Attention). On \texttt{TinyStories}, because \texttt{ShortConv1d(k=4)} aids local $n$-gram modeling at $4.9\text{M}$ tokens (helping both linear architectures outperform plain Softmax at $134.09$ PPL), the strictly controlled head-to-head is between the two linear architectures sharing identical convolutions and decay gates: 4D Stiefel-Tucker beats 2D Linear Attention by \textbf{$-4.98$ PPL} ($111.78$ vs.\ $116.76$) while using \textbf{9\% fewer backbone parameters} ($2.88\text{M}$ vs.\ $3.16\text{M}$).

\begin{figure}[!hb]
  \centering
  \includegraphics[width=0.88\linewidth]{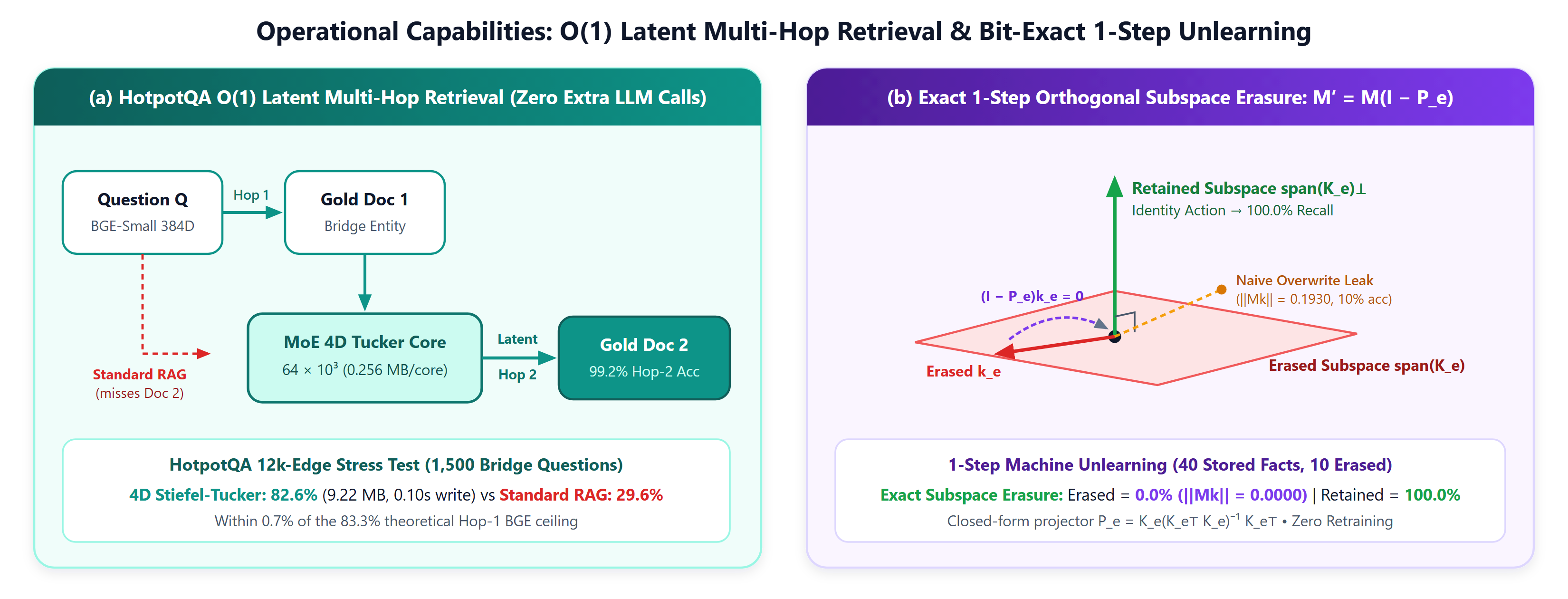}
  \caption{Operational capabilities of Stiefel-Projected 4D Tucker Memory. (a) On 12k-edge \texttt{HotpotQA}, latent Hop-2 routing through $0.256\text{ MB}$ cores achieves $99.2\%$ Hop-2 accuracy ($82.6\%$ Both-Gold@2 vs.\ $29.6\%$ for Standard RAG) with zero extra LLM calls. (b) Exact 1-step orthogonal subspace erasure $M' = M(I - P_e)$ collapses erased keys to bit-exact zero ($\|M'k\| = 0.0000$) while preserving $100.0\%$ recall on the orthogonal complement.}
  \label{fig:ops}
\end{figure}

\subsection{Exact 1-Step Orthogonal Subspace Erasure}

Finally, we validate Proposition~\ref{prop:erase} on a $D=64$ associative memory storing $40$ facts and erasing $10$ targets (Figure~\ref{fig:lm}(b) and Figure~\ref{fig:ops}). Naive 1-step zero-overwrite leaves substantial residual leakage ($10.0\%$ erased recall, $\|\mathbf{M}k\| = 0.1930$), whereas Exact Subspace Erasure drives erased recall to \textbf{0.0\%} with bit-exact zero norm (\texttt{||Mk||} $= 0.0000$) while preserving \textbf{100.0\%} recall on all $30$ retained facts.

\section{Discussion, Limitations, and Future Work}

Our results establish that lifting linear-attention states from 2D matrices to asymmetric, Stiefel-projected 4D Tucker cores breaks the degree-1 recall bottleneck without sacrificing $O(1)$ inference or requiring per-step SVD. Two honest limitations guide our next steps: (1) like Mamba and Gated DeltaNet, pure linear/cubic attention benefits from a short depthwise convolution (\texttt{ShortConv1d}) for local 1-token shifts in MQAR; and (2) our language-modeling validation is currently at 4-layer \texttt{TinyStories} scale ($4.9\text{M}$ tokens). Scaling 4D Stiefel-Tucker Attention to $100\text{M}+$ parameters over multi-billion-token corpora—and unifying it with our \texttt{QuatTucker-ZigZag} FFN backbones \citep{barmase2026quattucker} to create a fully tensor-factorized, $O(1)$-inference Transformer—is our immediate priority.

\bibliography{references}

@inproceedings{widrow1960adaptive,
  title={Adaptive switching circuits},
  author={Widrow, Bernard and Hoff, Marcian E.},
  booktitle={IRE WESCON Convention Record},
  pages={96--104},
  year={1960}
}

@article{tucker1966some,
  title={Some mathematical notes on three-mode factor analysis},
  author={Tucker, Ledyard R.},
  journal={Psychometrika},
  volume={31},
  number={3},
  pages={279--311},
  year={1966}
}

@article{hopfield1982neural,
  title={Neural networks and physical systems with emergent collective computational abilities},
  author={Hopfield, John J.},
  journal={Proceedings of the National Academy of Sciences},
  volume={79},
  number={8},
  pages={2554--2558},
  year={1982}
}

@article{smolensky1990tensor,
  title={Tensor product variable binding and the representation of symbolic structures in connectionist systems},
  author={Smolensky, Paul},
  journal={Artificial Intelligence},
  volume={46},
  number={1-2},
  pages={159--216},
  year={1990}
}

@article{schmidhuber1992learning,
  title={Learning to control fast-weight memories: An alternative to dynamic recurrent networks},
  author={Schmidhuber, J{\"u}rgen},
  journal={Neural Computation},
  volume={4},
  number={1},
  pages={131--139},
  year={1992}
}

@article{plate1995holographic,
  title={Holographic reduced representations},
  author={Plate, Tony A.},
  journal={IEEE Transactions on Neural Networks},
  volume={6},
  number={3},
  pages={623--641},
  year={1995}
}

@article{delathauwer2000multilinear,
  title={A multilinear singular value decomposition},
  author={De Lathauwer, Lieven and De Moor, Bart and Vandewalle, Joos},
  journal={SIAM Journal on Matrix Analysis and Applications},
  volume={21},
  number={4},
  pages={1253--1278},
  year={2000}
}

@inproceedings{vaswani2017attention,
  title={Attention is all you need},
  author={Vaswani, Ashish and Shazeer, Noam and Parmar, Niki and Uszkoreit, Jakob and Jones, Llion and Gomez, Aidan N. and Kaiser, {\L}ukasz and Polosukhin, Illia},
  booktitle={Advances in Neural Information Processing Systems (NeurIPS)},
  year={2017}
}

@inproceedings{yang2018hotpotqa,
  title={{HotpotQA}: A dataset for diverse, explainable multi-hop question answering},
  author={Yang, Zhilin and Qi, Peng and Zhang, Saizheng and Bengio, Yoshua and Cohen, William W. and Salakhutdinov, Ruslan and Manning, Christopher D.},
  booktitle={Proceedings of EMNLP},
  year={2018}
}

@inproceedings{katharopoulos2020transformers,
  title={Transformers are {RNNs}: Fast autoregressive transformers with linear attention},
  author={Katharopoulos, Angelos and Vyas, Apoorv and Pappas, Nikolaos and Fleuret, Fran{\c{c}}ois},
  booktitle={International Conference on Machine Learning (ICML)},
  year={2020}
}

@inproceedings{ramsauer2021hopfield,
  title={Hopfield networks is all you need},
  author={Ramsauer, Hubert and Sch{\"a}fl, Bernhard and Lehner, Johannes and Seidl, Philipp and Widrich, Michael and Adler, Thomas and Gruber, Lukas and Holzleitner, Markus and Pavlovi{\'c}, Milena and Sandve, Geir Kjetil and others},
  booktitle={International Conference on Learning Representations (ICLR)},
  year={2021}
}

@inproceedings{ravfogel2020null,
  title={Null it out: Guarding protected attributes by iterative nullspace projection},
  author={Ravfogel, Shauli and Elazar, Yanai and Gonen, Hila and Twiton, Michael and Goldberg, Yoav},
  booktitle={Proceedings of ACL},
  year={2020}
}

@inproceedings{schlag2021linear,
  title={Linear transformers are secretly fast weight programmers},
  author={Schlag, Imanol and Irie, Kazuki and Schmidhuber, J{\"u}rgen},
  booktitle={International Conference on Machine Learning (ICML)},
  year={2021}
}

@inproceedings{bourtoule2021machine,
  title={Machine unlearning},
  author={Bourtoule, Lucas and Chandrasekaran, Varun and Choquette-Choo, Christopher A. and Jia, Hengrui and Travers, Adelin and Zhang, Baiwu and Lie, David and Papernot, Nicolas},
  booktitle={IEEE Symposium on Security and Privacy (SP)},
  year={2021}
}

@article{eldan2023tinystories,
  title={{TinyStories}: How small can language models be and still speak coherent {English}?},
  author={Eldan, Ronen and Li, Yuanzhi},
  journal={arXiv preprint arXiv:2305.07759},
  year={2023}
}

@article{sun2023retentive,
  title={Retentive network: A successor to transformer for large language models},
  author={Sun, Yutao and Dong, Li and Huang, Shaohan and Ma, Shuming and Xia, Yuqing and Xue, Jilong and Wang, Jianyong and Wei, Furu},
  journal={arXiv preprint arXiv:2307.08621},
  year={2023}
}

@article{xiao2023cpack,
  title={{C-Pack}: Packaged resources to advance general {Chinese} embedding},
  author={Xiao, Shitao and Liu, Zheng and Zhang, Peitian and Muennighoff, Niklas},
  journal={arXiv preprint arXiv:2309.07597},
  year={2023}
}

@article{arora2023zoology,
  title={Zoology: Measuring and improving recall in efficient language models},
  author={Arora, Simran and Eyuboglu, Sabri and Timalsina, Aman and Johnson, Isys and Poli, Michael and Zou, James and Rudra, Atri and R{\'e}, Christopher},
  journal={arXiv preprint arXiv:2312.04927},
  year={2023}
}

@article{gu2023mamba,
  title={Mamba: Linear-time sequence modeling with selective state spaces},
  author={Gu, Albert and Dao, Tri},
  journal={arXiv preprint arXiv:2312.00752},
  year={2023}
}

@inproceedings{yang2024gla,
  title={Gated linear attention transformers with hardware-efficient training},
  author={Yang, Songlin and Wang, Bailin and Shen, Yikang and Panda, Rameswar and Kim, Yoon},
  booktitle={International Conference on Machine Learning (ICML)},
  year={2024}
}

@inproceedings{dao2024transformers,
  title={Transformers are {SSMs}: Generalized models and efficient algorithms through structured state space duality},
  author={Dao, Tri and Gu, Albert},
  booktitle={International Conference on Machine Learning (ICML)},
  year={2024}
}

@article{yang2024gateddeltanet,
  title={Gated delta networks: Improving {Mamba2} with delta rule},
  author={Yang, Songlin and Kautz, Jan and Hatamizadeh, Ali},
  journal={arXiv preprint arXiv:2412.06464},
  year={2024}
}

@article{behrouz2024titans,
  title={Titans: Learning to memorize at test time},
  author={Behrouz, Ali and Zhong, Peilin and Mirrokni, Vahab},
  journal={arXiv preprint arXiv:2501.00663},
  year={2025}
}

@article{barmase2026quattucker,
  title={Quaternion-{Tucker} {ZigZag} networks: Halving transformer backbone parameters and enabling exact {2D} zero-loss continual scaling},
  author={Barmase, Shubham},
  journal={arXiv preprint arXiv:2611.33983},
  year={2026}
}

\end{document}